\pdfoutput=1

\documentclass[11pt]{article}

\usepackage[]{acl}
\usepackage{graphicx}
\usepackage{times}
\usepackage{amsmath}
\usepackage{latexsym}
\usepackage{multirow}
\usepackage{float}
\usepackage{booktabs}
\usepackage{array}
\newcolumntype{L}[1]{>{\raggedright\let\newline\\\arraybackslash\hspace{0pt}}m{#1}}
\newcolumntype{C}[1]{>{\centering\let\newline\\\arraybackslash\hspace{0pt}}m{#1}}
\newcolumntype{R}[1]{>{\raggedleft\let\newline\\\arraybackslash\hspace{0pt}}m{#1}}
\usepackage[T1]{fontenc}

\usepackage[utf8]{inputenc}

\usepackage{microtype}

\usepackage{graphicx}
\usepackage{booktabs}
\usepackage{cellspace}
\graphicspath{ {./image/} }
\usepackage{pgf-pie}

\usepackage{todonotes}

\usepackage{enumitem}
\usepackage{amssymb}

\title{MBTI Personality Prediction for Fictional Characters Using Movie Scripts}

\author{%
  Yisi Sang$^{1}$\thanks{\,\,Authors contributed equally to this paper. Mo Yu is the corresponding author.} \quad Xiangyang Mou$^{2*}$ \quad Mo Yu$^{3*}$ \quad Dakuo Wang$^{4}$ \quad Jing Li$^{5}$ \quad  Jeffrey Stanton$^{1}$ \\
  \small{$^1$Syracuse University \qquad $^2$Rensselaer Polytechnic Institute \qquad $^3$Pattern Recognition Center, WeChat AI} \\ \small{$^4$IBM Research, Northeastern University \qquad $^5$New Jersey Institute of Technology} \\ 
  \small{\texttt{yisang@syr.edu}\quad \texttt{moux4@rpi.edu} \quad \texttt{moyumyu@tencent.com}}}

\begin{document}
\maketitle
\begin{abstract}
An NLP model that understands stories should be able to understand the characters in them.
To support the development of neural models for this purpose, we construct a benchmark, \texttt{Story2Personality}.
The task is to predict a movie character's MBTI or Big 5 personality types based on the narratives of the character.
Experiments show that our task is challenging for the existing text classification models, as none is able to largely outperform random guesses.
We further proposed a multi-view model for personality prediction using both verbal and non-verbal descriptions, which gives improvement compared to using only verbal descriptions. The uniqueness and challenges in our dataset call for the development of narrative comprehension techniques from the perspective of understanding characters.\footnote{Our code and data are released at \url{https://github.com/YisiSang/Story2Personality}}

\end{abstract}

\section{Introduction}


Character comprehension is commonly regarded as the cornerstone to comprehending stories in psychology and education~\cite{bower1990mental,paris2003assessing,zhao2022educational}. The NLP community has done some work on character comprehension in reading comprehension tasks, but most of the existing studies focus on short or expository texts (e.g., story summaries)~\cite{urbanek2019learning,brahman2021let}.
Moreover, most of them are limited in factoid understanding of characters, such as coreference resolution~\cite{chen2016character} and character relationships~\cite{iyyer2016feuding}, and
few studies have explored deeper comprehension of characters' persona~\cite{flekova2015personality, sang2022survey}, on which humans can generally do well.
\begin{figure}[t!]
 \centering
 \includegraphics[width=0.85\columnwidth]{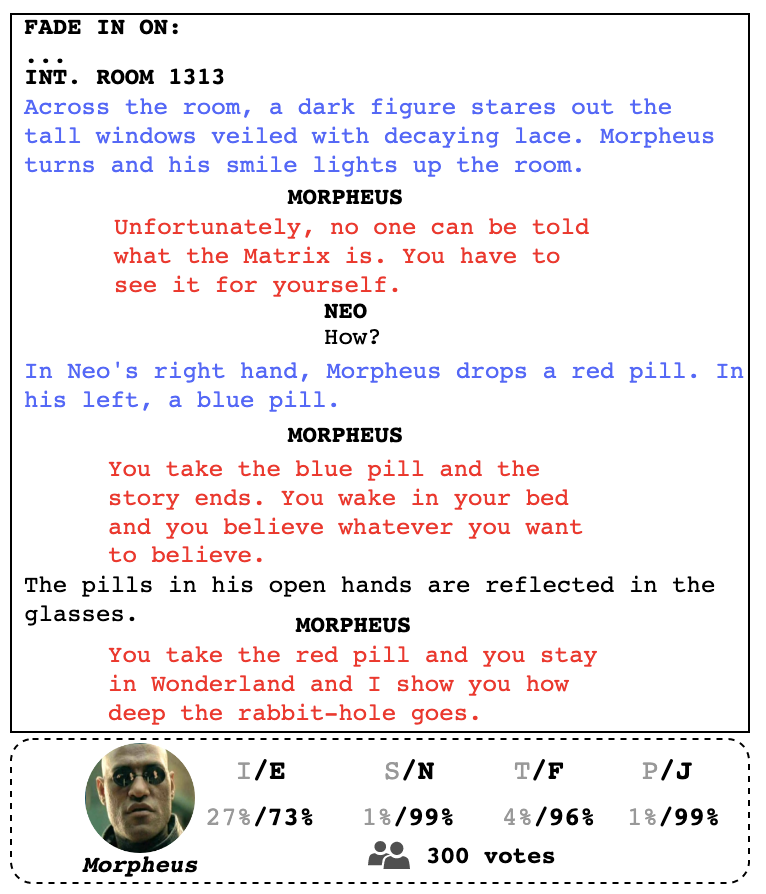}
 \vspace{-10pt}
 \caption{\small{An example excerpt from the movie script of ``The Matrix''. \textcolor{blue}{Blue} utterances are the character \textit{Morpheus's} \textbf{scene descriptions}, \textcolor{red}{red} are his \textbf{dialogues}.  \textit{Morpheus's} MBTI personality was rated as \textbf{ENFJ} by 300 user votes.}}
 \label{fig:matrix}
 \vspace{-15pt}
\end{figure}
We propose \texttt{Story2Personality}, a new narrative understanding benchmark to encourage the study of character understanding. 
The goal of \texttt{Story2Personality} is to predict personality according to the character's narrative texts in the script. 
%
%

Personality prediction from narratives has many challenges. First, stories often use a variety of narrative clues (e.g., scenery changes), sequence (e.g., flashback) and rhetorical techniques (e.g., metaphor)~\cite{xu-etal-2022-fantastic}.
%
%
%
Second, the inputs of the task are long ($>$10K words on average), challenging the applications of Transformer-based models~\cite{vaswani2017attention}. Third, both the scene descriptions and dialogues are informative for the prediction, requiring models to jointly consider multiple views of inputs.




This study makes the following contributions:
\vspace{-5pt}
\begin{itemize}[leftmargin=*]
\setlength\itemsep{0em}
    \item We establish a large-scale dataset for personality prediction of narrative characters that can support the development of neural models. Our dataset consists of 3,543 characters from 507 movies with MBTI labels of four dimensions.
    In comparison, the only existing related dataset~\cite{flekova2015personality} contains only 298 characters and focuses on a single dimension. 
    Our dataset is proved challenging --- on this binary classification task, none of the baselines achieve higher than 60\% macro-F1.
    \vspace{-5pt}
    \item We develop a movie script parser to automatically process a script to a structured form with the verbal character dialogues and the non-verbal scene descriptions illustrating backgrounds.
    Human study shows that our parser is more accurate compared to previous rule-based tools.
    \vspace{-5pt}
    \item We propose an extension to BERT classifier~\cite{devlin2018bert}  to handle the long and multi-view (verbal and non-verbal) inputs. Our model improves 2-3\% over the baselines. This shows the potential of exploiting both verbal and non-verbal narratives of characters, which is consistent with psychological theory~\cite{mccroskey1996fundamentals,richmond2008nonverbal}; and suggests directions of future model design.
\end{itemize}

\section{Related Work}
\label{app:relatework}

\paragraph{Character-Centric Narrative Understanding}
There have been existing studies on character-centric narrative understanding.
While many of them \cite{massey2015annotating,srivastava2016inferring,brahman2021let}
work on summaries of stories or summaries of characters.
Their scopes thus have a different assessment purpose from ours, and have the challenge on understanding long narrative inputs greatly reduced.

For works that use long narratives, most of them study the inter-character relationship~\cite{elson2010extracting,elsner2012character,elangovan2015you,iyyer2016feuding,chaturvedi2016modeling,chaturvedi2017unsupervised,kim2019frowning}. Inter-character relationship is also related to social network theories. Various of relationships have been considered in these studies, while most of them rely on unsupervised learning and do not provide labeled data for a direct automatic evaluation. TVSHOWGUESS explored multiple perspectives of persona using long narratives but the task format is different from us~\cite{sang2022tvshowguess}.

Finally, there is work on fundamental NLP annotating techniques over books and screenplays, such as named entity recognition~\cite{bamman2019annotated}, coreference resolution~\cite{chen2016character}, event-centric extraction~\cite{xu2022nece}, and entity-centric natural language modeling~\cite{clark2018neural} which is different from narrative understanding. Their techniques can be helpful to our task but the scope of their research is different from character-centric comprehension.

\paragraph{Latent Persona Induction}
Besides \cite{flekova2015personality} that is similar to our work in terms of the focus on personality classification, there is another line of related work on latent persona induction~\cite{bamman2013learning}. The work learns a topic model over character behaviors from books, and each latent topic corresponds to an induced persona. The induced persona vectors can be then applied to potential applications as a type of character representation.

From the perspective of practicality, our work and \cite{bamman2013learning} have their own strengths. From our motivation of story comprehension assessment, the difference is whether we provide a direct evaluation of the character understanding or evaluate it in down-streaming tasks – similar to the aforementioned relationship detection work, it is also difficult to provide an automatic and objective evaluation for the task of \cite{bamman2013learning}. 
The advantage of our task is that it supports direct automatic evaluation by itself, without the need for further downstream tasks; and it can be also used to evaluate the methods for the task of \cite{bamman2013learning}. Moreover, compared to a direct evaluation, the performance on a down-streaming task can be affected by other factors other than persona so a good performance on downstream tasks may not come directly caused by a good persona representation. 
The cons of our task is that it is limited to the personality types that have human annotations.

\section{Background of MBTI}
Personality is a ``stable and measurable'' individual characteristic~\cite{vinciarelli2014survey} which can ``distinguish internal properties of the person from overt behaviors''~\cite{matthews2003personality}. Understanding the personalities of the characters is essential for grasping the story's greater message.
%
%
The Myers–Briggs Type Indicator (MBTI)~\cite{myers1962myers} and the Big-5 Personality are two of the most popular personality scales. We used MBTI as the annotation criteria since despite some validity controversy in self-report measurement, research shows that a person’s friend can accurately judge his/her MBTI personality~\cite{cohen1981construct}. In our narrative comprehension scenario, a fictional character’s MBTI personality is judged by other human raters in an online community, which is quite similar to \textbf{the third-person evaluation} scenario, and should yield a reasonable validity. We also conducted our study on Big-5 and reported the results in Appendix ~\ref{sec:appendixBig5}.



MBTI assess the psychological preferences in how people perceive the world and make decisions in four dimensions: 
E/I: extravert (E) is seen as being generally active and objective while the intravert (I) is seen as generally passive and subjective~\cite{sipps1987multifactorial}. S/N: sensing (S) is seen as attending to sensory stimuli; intuition (N) describes a more detached, insightful analysis of events and stimuli~\cite{boyle1995myers}. T/F: thinking (T) involves logical reasoning and decision making; feeling (F) involves a more subjective and interpersonal approach~\cite{thomas1983field}. J/P: judging (J) attitude is associated with prompt decision making; perception (P) involves greater patience and waiting for more information before making a decision. An individual's MBTI type has a label based on her dominant preference for each dimension.
In Figure \ref{fig:matrix}, Morpheus is an extraversion person, understanding the world with intuition, dealing with things with feeling, and organize the world around him by judging. Together gives an ENFJ type.

\section{\texttt{Story2Personality} Dataset}
We constructed our dataset in three stages: extracting movie scripts from the Internet Movie Script Database (\textit{IMSDB}~\footnote{\url{https://imsdb.com/}}), parsing the collected movie scripts into dialogue and scene sections, matching characters' personality types from The Personality Database(\textit{PDB}\footnote{\url{https://www.personality-database.com/}}) with their dialogues and scenes.

\subsection{Movie Scripts Collection}
We collected HTML files from IMSDB combined with movie scripts in NarrativeQA~\cite{kovcisky2018narrativeqa}. After removing corrupted or empty files, we got 1,464 usable scripts. 

 


\subsection{Our Statistical Movie Script Parser}
\label{ssec:script_parser}

As shown in Figure~\ref{fig:matrix}, a movie script usually has four basic format elements~\cite{riley2009hollywood}: \textbf{Scene Headings}, one line description of each scene's type, location, and time (i.e., \texttt{INT. ROOM 1313}); \textbf{Scene description}, the description of the actions of the characters (i.e., text in \textcolor{blue}{blue}); \textbf{Dialogues}, names of characters and actual words they speak (i.e., text in \textcolor{red}{red}); \textbf{Transitions}, instructions for linking scenes together (i.e., \texttt{FADE IN ON}). 

In order to extract dialogues and scene descriptions in a structured form, we first split the scripts to sections, i.e., text chunks between two adjacent bolded chunks which are scene headings or character names and stored the bolded texts as section titles.
Then we designed a statistical method to classify the section types:


\medskip
\noindent\textbf{Rule-Based Pre-Processing }
We start with a rule to classify the sections into dialogues and scenes.
As Figure~\ref{fig:matrix} shows, a common format of movie scripts is to align the shot headings, transitions and scene descriptions vertically, and uses a larger indentat for dialogues. 
So, the indent size can be used to identify dialogues.
Since the indentat size may vary across different scripts.
Our rule assumes the sections as dialogues if they have larger indent compared to \texttt{FADE IN} in the same script and the others as scenes.

\medskip
\noindent\textbf{Silver Parses Construction }
The rule-based pre-processing introduces many noises.
We then designed a statistical method to automatically determine the threshold indent of dialogues.
First, we compute the averaged ratio $\mu$ of dialogues in a script and its standard variation $\sigma$.
Second, we keep adding sections with the largest indent sizes to the set of dialogues, 
until the ratio of added sections becomes larger than $\mu$$+$$\sigma$.
Finally, we keep the left sections as scenes. If none of the indentation size can reach the ratio of dialogues in the range of $\mu$$\pm$$\sigma$, the movie script was seen as a failure case.
We designated the successfully processed scripts with the dialogues/scene labels as the ``silver'' set which consists of 29\% of the scripts.




\medskip
\noindent\textbf{Section Classifier }
For the failure scripts from the previous step and the scripts without \texttt{FADE IN} markers, we 
trained a BERT-based section classifier using 137,042 labeled sections from the silver set to label them.
%
%
The classifier achieved 99.31\% accuracy on a held out validation set. 
The outputs are our final parses.
 
\subsection{Personality Collection and Mapping} 
We collect human rated MBTI types from \emph{PDB}. Movie scripts are the blueprint for the actor’s performance. An actor’s body language, dialogue, and contexts are all described in the scripts~\cite{jhala2008exploiting}. Human rater's perception of a character’s personality from the movies would be consistent with the script’s description.
In total, we collected MBTI types of 28,653 characters. Each character has an id, name, vote count, and voters' agreement on each MBTI dimension. 
For example, the MBTI profile in Figure~\ref{fig:matrix} has 300 voters, with different agreement rate along each dimension.
%
%
To ensure the quality of personality voting, we removed character profiles with $<$3 voters and $<$60\% agreement rate so some characters do not have all the 4 dimensions. 
We include more details in Table \ref{tab:16type} in the appendix. \textit{PDB's} When the user starts rating, the rating interface hides the previous rater's choices. Thus, the rater would not have prior bias. We then matched the characters' personality profiles to the scripts, if the name can be softly matched to the dialogue title or the recognized named entities in the scenes (details and example of the final processed data in Appendix~\ref{sec:appendix}). Table \ref{tab:4dimension} shows the core statistics of our dataset. The numbers of data points can also be found in Table \ref{tab:agreement}.

\begin{table}[t]
\scriptsize
\centering
\small
\begin{tabular}{lp{16mm}ccc}
\toprule
& Dimension & Train(\%) & Dev(\%) &Test(\%)  \\ \midrule
\multirow{4}{*}{\rotatebox[origin=c]{90}{(a)}} 
&E/I    & 45.9/51.8 & 49.6/49.0 & 52.6/44.2     \\ 
&N/S    & 36.6/60.4 & 41.8/54.0 & 41.4/55.0     \\
&T/F    & 54.7/43.2 & 45.8/50.8 & 46.0/52.8     \\
&J/P    & 46.4/51.3 & 47.2/51.2 & 45.6/53.0     \\
\midrule
\end{tabular}

\begin{tabular}{llccc} 
\midrule
&           &  Mean     & Min       &  Max          \\ 
\midrule    \multirow{4}{*}{\rotatebox[origin=c]{90}{(b)}}
& \# dialogues/character & 76.90     & 0 & 776       \\
& \# words/dialogue     & 917.74    & 1 & 12, 536   \\
& \# scenes/character    & 41.08     & 0 & 495       \\
& \# words/scene        & 1,381.47  & 1 & 25,457    \\
\bottomrule
\end{tabular}
\vspace{-7pt}
\caption{Distribution of two personality types per dimension (a) and core statistics (b) in \texttt{Story2Personality}.}
\label{tab:4dimension}
\end{table}

\vspace{-7pt}
\section{Dataset Analysis}
We conduct human study to verify the advantage of our script parser; then provide the human performance on our dataset.

\medskip
\noindent\textbf{Script Parsing results } 
We compared our parsing results with the results of the state-of-the-art open-sourced script parser~\cite{ramakrishna2017linguistic}, which employs many human written rules, with a human study.
We randomly selected five scene descriptions and five dialogue sections in 10 common movies, giving 100 snippets for evaluation (40 from the silver set).
Then we manually compared the parsing results with the original movie scripts. Table~\ref{tab:parsing} shows the results.
Our parser outperforms \citet{ramakrishna2017linguistic} with a large margin. Most mistakes of \cite{ramakrishna2017linguistic} is to recognize scenes as dialogues. There are other parsers but did not publish the code or data, so we cannot conduct human study for comparison. A state-of-the-art learning model~\cite{agarwal2014parsing} reports 91\% accuracy on line-level classification. In a preliminary study, we achieve 99\% on this task, but finally choose do conduct more accurate section-level classification as in Section \ref{ssec:script_parser}.


\begin{table}
\centering
\resizebox{\columnwidth}{!}{%
\begin{tabular}{lrr} 
\toprule                                                                                            
    & Correct scene        & Correct dialogue        \\ \midrule
\citet{ramakrishna2017linguistic}  &85\%  &93\% \\
Our parser  &\bf 97\%  &\bf 100\% \\

\bottomrule
\end{tabular}
}
\vspace{-5pt}
\caption{\small{Comparison of correct parsing results.}}
\vspace{-10pt}
\label{tab:parsing}
\end{table}

\medskip
\noindent\textbf{Human performance }
We take the majority vote of each character's MBTI types as the groundtruth.
This gives an averaged {93.54\%} human accuracy across the four personality dimensions on our test data.
Computing humans' macro-F1 score lacks an analytical form from the agreement scores. Therefore we make an approximation by sampling three voters (the minimum number of voters in our dataset) for each character and treating them like the predictions of three different models. This gives overall $>$95\% scores which is much higher than model performance (in Table~\ref{tab:performance}).
The statistics of human agreement on MBTI dimensions is shown in Table \ref{tab:agreement} in the appendix.

\begin{figure}
\centering
\includegraphics[width=60mm,scale=0.6]{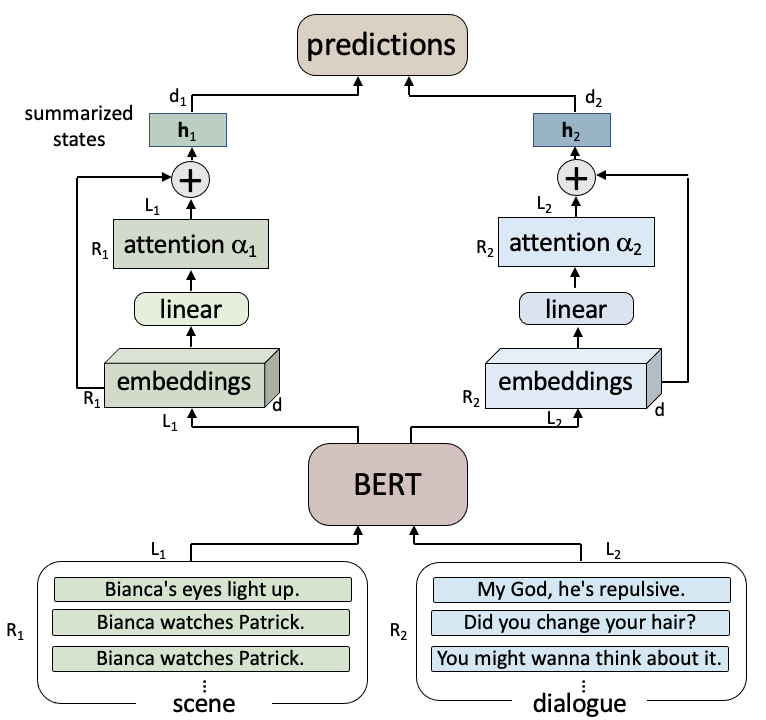}
\vspace{-10pt}
\caption{MR-MV BERT model architecture.}
\centering
\label{fig:MVMR_BERT_arch}
\vspace{-14pt}
\end{figure}

\section{Experiments}
\paragraph{Baselines}
We build two baseline models.
$\bullet$\textbf{SVM}, the LinearSVC from sklearn.svm. We extracted top 20K word unigram, bigram, and trigram features according to term frequency after removing stop words. We set $C$$=$0.1. $\bullet$\textbf{ BERT},
fine-tuning the out-of-box BERT, with a linear head on the `[CLS]' token's final layer embedding for classification.


\paragraph{Our Method} We propose the multi-view multi-row BERT (\textbf{MV-MR BERT}) classifier (Fig.~\ref{fig:MVMR_BERT_arch}) which is an extension of BERT to deal with the long inputs and handle the verbal and non-verbal information differently.
%
First, to handle the long input per character, we borrow the idea from fusion-in-decoder~\cite{izacard2020leveraging}. Since the complexity of Transformers is $O(RL^2)$ (with $R$ the number of rows and $L$ the length per row), when $L$ is very large, we can split it into multiple segments to reduce the quadratic term. Next, we rely on the attention over all the segments to fuse the information.
Specifically, we split the input content $\mathcal{D}$ of a character into multiple segments $\mathcal{D} = \{\mathcal{S}_i\}^R_{i=1}$, and encode all the segments in a minibatch as
$\mathbf{H}=\text{BERT}(\mathcal{S}_i) \in \mathbb{R}^{R \times L \times d}$, where 
$d$ is the hidden state size.
Then a linear head is applied to get the attention score across tokens in all the rows as $\mathbf{\alpha} = \text{softmax}(\mathbf{H} W + b) \in [0,1]^{R\times L}$.
The final summarized representation of the input $\mathcal{D}$ is thus the weighted summation $\mathbf{h}_{\mathcal{D}} = \sum_{i=1}^R \sum_{j=1}^L \alpha_{ij}\mathbf{H}_{ij}$.
Second, to handle both the dialogue and behavioral description a character,
%
%
our multi-view model receives an input pair $(\mathcal{D}^{\text{dial}}, \mathcal{D}^{\text{scene}})$, then uses a shared BERT and separated linear heads 
to compute the 
summarized states $\mathbf{h}_{\mathcal{D}}^{\text{dial}}$ and $\mathbf{h}_{\mathcal{D}}^{\text{scene}}$. The two vectors are fed into a fully-connected layer for prediction.
For the scene descriptions, we prepend a special token ``[ent]'' to the target character's name to denote its position. The attention $\alpha^{\text{scene}}$ is only computed on these special tokens.





\paragraph{Results and Analysis}
Following~\citet{flekova2015personality}, we use macro-averaged F1 as evaluation metric.
%
Table \ref{tab:performance} shows the main results on the four MBTI dimensions. Peak performance was achieved by our MV-MR BERT. The result suggests using both dialog and action scene descriptions consistently improved model performance.

The results are generally low compared to human performance, showing the task is challenging to existing models. We analyzed the learning curve of BERT model by adding the training data from 1K to 2.5K characters (Table~\ref{tab:learning_curv} in Appendix~\ref{sec:appendixModel}). The model performance did not change a lot in the development dataset.
Figure~\ref{fig:BERT+MVMR-BERT} in Appendix~\ref{sec:appendixModel} gives further evidence for the challenge of our task, which
shows the dev and test results are not highly-correlated, meaning that by achieving near perfect accuracy on the training data, the models largely overfit the noises instead of capturing real clues.







\begin{table}[t!]
\resizebox{\columnwidth}{!}{%
\begin{tabular}{@{}llllll@{}} 
\toprule
Model & E/I & N/S & T/F & J/P \\\midrule

SVM & 54.65 & 55.41& 52.83  & 56.18 \\  
BERT  & 56.06$_{\pm\textrm{0.73}}$& 55.59$_{\pm\textrm{3.36}}$& 57.13$_{\pm\textrm{0.97}}$&57.59$_{\pm\textrm{1.40}}$\\
MV-MR BERT & \bf 57.50$_{\pm\textrm{2.04}}$ & \bf 57.42$_{\pm\textrm{4.27}}$& \bf 60.33$_{\pm\textrm{0.93}}$& \bf 59.83$_{\pm\textrm{1.42}}$ \\
\quad - multiview &57.30$_{\pm\textrm{1.91}}$& 57.05$_{\pm\textrm{1.80}}$& 57.04$_{\pm\textrm{2.05}}$& 57.39$_{\pm\textrm{2.21}}$\\ 
\midrule
Human Perf. &98.19$_{\pm\textrm{0.60}}$&97.82$_{\pm\textrm{0.10}}$&98.51$_{\pm\textrm{0.67}}$&98.03$_{\pm\textrm{0.19}}$\\
\bottomrule
\end{tabular}
}
\vspace{-5pt}
\caption{Macro F1 scores on the four dimensions}
\label{tab:performance}
\vspace{-15pt}
\end{table}


\paragraph{Model Performance on the Big Five Personality Test}
\label{sec:appendixBig5}
We collected a variation of Big 5, the Global 5, from the PDB. The Global 5 adaptation of the Big Five~\cite{digman1997higher} consists of Extroversion, Emotional Stability, Orderliness, Accommodation, and Intellect. The SLOAN nation system is the scoring format for Global 5 test. SLOAN nation keys are: \underline{\textbf{S}}ocial/ \textbf{R}eserved, \underline{\textbf{L}}imbic/ \textbf{C}alm, \underline{\textbf{O}}rganized/ \textbf{U}nstructured, \underline{\textbf{A}}ccommodating/ \textbf{E}gocentric, \underline{\textbf{N}}on-curious/ \textbf{I}nquisitive. The number of people who evaluated the movie characters with big5 was very small, and after eliminating the characters without dialogues, we only got 1,346 characters' data. Such a small amount of data deeply affects the training of neural-based model. Table~\ref{tab:big5} shows the model performance.

\begin{table}[h]
\small
\centering
\begin{tabular}{@{}lllllll@{}} 
\toprule
Model & S/R & L/C & O/U & A/E & N/I \\\midrule

SVM &57.31&55.93&59.40&53.51&55.89 \\  
MV-MR BERT & 58.98 & 54.20 & 62.26& 54.00 & 59.04\\

\bottomrule
\end{tabular}
\vspace{-5pt}
\caption{Macro F1 scores on the Big 5 dimensions}
\label{tab:big5}
\end{table}

\vspace{-5pt}
\section{Limitation}
Movie scripts are the blueprint for the actor’s performance. An actor’s dialogue, body language, and the contexts are well described in the scripts. There is sufficient information in the scripts for readers to understand the characters. However, the actors' portrayal have the potential to add additional cues to influence the audience's perception of the fictional characters' personalities. In future work we will try to use multi-modality data as input.

\section{Conclusion}
We develop a movie script parser and proposed a new narrative understanding benchmark, \texttt{Story2Personality}, which enables neural model training for understanding characters.
We evaluate several classifiers on our task -- while our multi-view multi-view BERT model achieves a substantial improvement over the SVM and BERT baselines, there is a huge gap compared to human performance. 
This indicates our dataset a valuable and challenging task for future research. In the future we will expand our dataset.

\clearpage
\newpage

\bibliographystyle{acl_natbib}
\bibliography{custom} 

\clearpage
\newpage

\appendix

\section{Details of Dataset Construction}
\paragraph{Soft Name Matching Algorithm}
We created two movie-character dictionaries to associate the characters with the movies using the characters' full names and their subcategories (i.e., movie names) in personality profile data, as well as section titles (i.e., character names or scene headings) and movie names in the movie script data. Then, we tokenized and lowercased the character names. We matched both the exact same full names and the intersections of tokens such as the first or last name of the full name when the movie names are matched. To identify a character's scene descriptions, we extracted named entities from scene descriptions and then matched the characters and scenes based on their names using the same method. After matching the character name with the movie name, we store the MBTI personality, vote count, dialogue and scene descriptions into a dictionary for each character. 

\begin{figure*}[ht]
\centering
\includegraphics[width=0.9\textwidth]{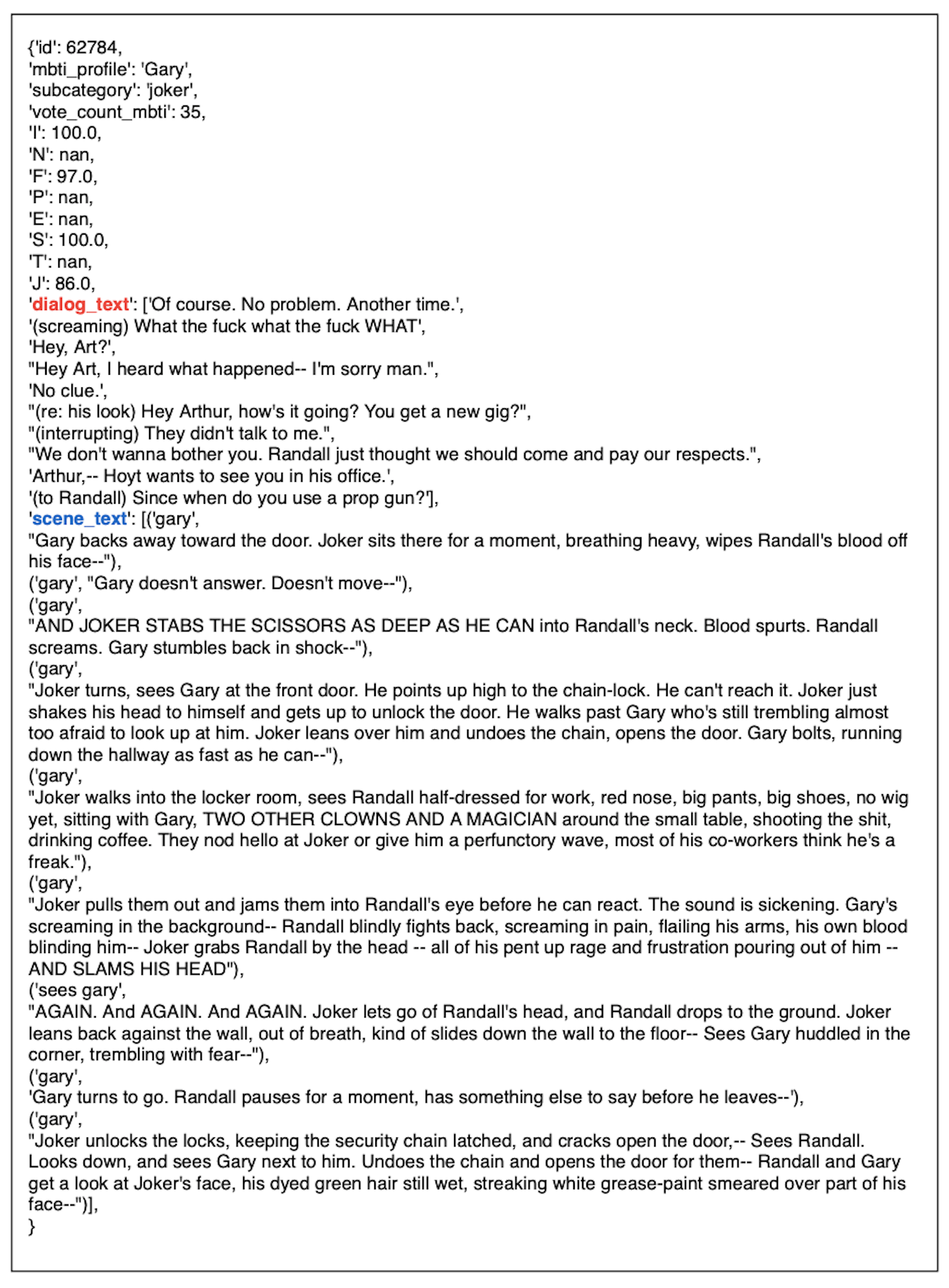}
\caption{\small{An example except from \texttt{Story2Personality}. Each character has ID ('id'), character name ('mbti\_profile'), movie name ('subcategory'), dialogue, and scene descriptions.}
\label{fig:character_example}}
\end{figure*}

\paragraph{Example Data Item for One Character}
Figure \ref{fig:character_example} shows the example of information for one character \textit{Gary} from the movie ``Joker'' in our \texttt{Story2Personality}, stored in json format.
The data item contains the ID (`id'), character name (`mbti\_profile') and movie name (`subcategory') in the PDB website; together with the human voted MBTI types and the number of votes.
Finally we save the dialogues of the character and the scenes he appears in two separated entries. For scenes, we save both the scene texts and the soft matched name mentions in the texts for the target character. The name mention is used to prepend the special tokens in our MV-MR BERT model.




\label{sec:appendix}


\section{Additional Information of the Dataset}
Table~\ref{tab:16type} lists the distribution of all the 16 MBTI types in our dataset, together with a representative movie character for each type.

\begin{table}[h]
\centering
\resizebox{\columnwidth}{!}{%
\begin{tabular}{@{}lll@{}} 
\toprule
Personality & \%  & Example \\ \midrule
ISTJ & 8.41\%       & \textit{Darth Vader} (``Star Wars'')\\
ISTP & 8.07\%       & \textit{Shrek} (``Shrek'')\\
ESTP & 8.21\%       & \textit{Han Solo} (``Star Wars'')       \\
ESTJ & 6.52\%       & \textit{Boromir} (``The Lord of the Rings'')\\
ISFJ & 6.41\%       & \textit{Forrest Gump} (``Forrest Gump'')  \\
ISFP & 6.49\%       & \textit{Harry Potter} (``Harry Potter'')\\
ESFJ & 4.88\%       & \textit{Cher Horowitz} (``Clueless'')\\
ESFP & 7.06\%       & \textit{Jack Dawson} (``Titanic'')\\
INFJ & 4.80\%       & \textit{Edward Cullen} (``Twilight'')\\
INFP & 5.42\%       & \textit{Amélie Poulain} (``Amélie'')\\
ENFP & 3.90\%       & \textit{Anna} (``Frozen'') \\
ENFJ & 3.75\%       & \textit{Judy Hopps} (``Zootopia'')       \\
INTJ & 4.26\%       & \textit{Michael Corleone} (``The God Father'')\\
INTP & 3.75\%       & \textit{Neo} (``The Matrix'')        \\
ENTP & 4.94\%       & \textit{Tyler Durden} (``Fight Club'') \\
ENTJ & 4.88\%       & \textit{Patrick Bateman} (``American Psycho'')    \\
 
\bottomrule
\end{tabular}
}
\caption{Distribution of the 16 MBTI personality types in \texttt{Story2Personality}}
\label{tab:16type}
\end{table}

\section{Statistics of Human Agreement}
Table~\ref{tab:agreement} lists the human agreement score on each MBTI dimension, on which we compute the human accuracy and approximate the human macro-F1 scores.

The raters are most divided in annotation of N, with an average agreement is 91.06\%
and the standard deviation 0.11.
One reason is that the perceptual style dimension N/S measures how the individual obtain information. Comparing with dimensions related to attitudes (E/I) or decision making (T/F, J/P)~\cite{jung2016psychological} perceptual style is more implicit. Specifically, S is seen as attending to sensory stimuli, while N describes a more detached, insightful analysis of events and stimuli~\cite{boyle1995myers}. 
They are more difficult to determine from the explicit story narratives.
\begin{table}[h]
\small
\centering
\begin{tabular}{lrrrrr} 
\toprule                                                                                            
    & Mean        & Min       & Max  & STD & \#Character  \\ \midrule
I  &94.43\%  &60\% &100\% & 0.10& 1,783\\
E  &94.22\% &60\% &100\% & 0.10&1,679\\
N  &91.06\%  &60\% &100\% &0.11&1,347\\
S  &93.32\%&60\%  &100\% &0.11&2,082\\
T  &94.22\%  &60\% &100\% &0.10&1,851\\
F  &93.68\% &60\% &100\% &0.10&1,617\\
P  &93.68\% & 60\% &100\% &0.10&1,825\\                
J  &93.72\% & 60\%    &100\%      &   0.10&1,644 \\                
             
\bottomrule
\end{tabular}
\caption{Descriptive statistics of voters' agreement}
\label{tab:agreement}
\end{table}

             

        

\begin{figure}
\centering
\vspace{-10pt}
\includegraphics[width=0.48\textwidth]{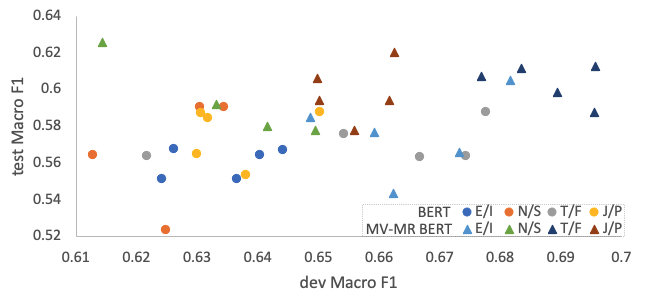}
\vspace{-20pt}
\caption{\small{Dev vs. test F1 scores of BERT-based models.}}
\centering
\label{fig:BERT+MVMR-BERT}
\vspace{-10pt}
\end{figure}

\section{Additional Model Performance}
\label{sec:appendixModel}
Figure~\ref{fig:BERT+MVMR-BERT} gives further evidence for the challenge of our task, which
plots the dev versus test scores during our model selection. It shows the dev and test results are not highly-correlated, meaning that by achieving near perfect accuracy on the training data, the models largely overfit the noises instead of capturing real clues. Both length and multiview have an improvement on model performance, but length has a slightly smaller impact, as shown in Table~\ref{tab:ablation}, when increasing the number of input tokens, the performance is not greatly affected.

\begin{table}[t]
\small
\centering
\begin{tabular}{lrrrr} 
\toprule                                                                                            
    & 1K      &1.5K  & 2K    & $\sim$2.5K    \\ \midrule
SVM  & 50.33 & 52.19 & 54.56& 55.41 \\
BERT  & 54.32 &55.42 &55.58&55.59 \\
\bottomrule
\end{tabular}
\vspace{-5pt}
\caption{\small{Learning curve with varying amount of training data (on N/S).}}
\vspace{-5pt}
\label{tab:learning_curv}
\end{table}

\begin{table}[h]
\small
\centering
\begin{tabular}{lrrrr} 
\toprule                                                                                            
    & 0.5K      &1K  & 2K &4K     \\ \midrule
J/P  &59.79  &58.18 &60.35&58.77 \\
T/F &59.91  &64.31 &63.32&65.42 \\
N/S  &56.15  &53.86 &55.65&57.18 \\
I/E  &61.64  &61.05 &62.87&62.69 \\
\bottomrule
\end{tabular}
\vspace{-5pt}
\caption{\small{{Ablation experiment on input length.}}}
\vspace{-5pt}
\label{tab:ablation}
\end{table}

\section{Model Checklist}
\label{app:checklist}

We implement our baselines based on HuggingFace Transformers.\footnote{https://github.com/huggingface/transformers} We use the pre-trained \texttt{bert-base-uncased} models.
We train all the models with the Adam optimizer.

We train our model on a single V100 GPU.
It takes around 2 hour and 10 minutes to train a multi-row BERT model.
For all the models, we train in total 20 epochs. 
\paragraph{Hyperparameters}
We set the number of rows in MV-MR BERT to 20, to maximize the usage of GPU memory. We set the learning rate to 2e-5. We report the average performance of five runs. 

\end{document}